\documentclass[runningheads]{llncs}
\usepackage{booktabs}
\usepackage{xcolor}
\usepackage{mathtools}
\usepackage{subcaption}
\usepackage{graphicx}
\usepackage[final]{listings}
\usepackage{float}
\usepackage{wrapfig}

\DeclarePairedDelimiterX{\norm}[1]{\lVert}{\rVert}{#1}

\newcommand{\argmin}[1]{\underset{#1}{\operatorname{arg}\,\operatorname{min}}\;}

\usepackage{adjustbox}
\usepackage{array}
\usepackage{hyperref}

\makeatletter
\newcommand{\printfnsymbol}[1]{%
  \textsuperscript{\@fnsymbol{#1}}%
}
\makeatother

\begin{document}

\title{TSInsight: A local-global attribution framework for interpretability in time-series data}
%
%
\author{Shoaib Ahmed Siddiqui\inst{1}\inst{2}\thanks{Equal contribution}\orcidID{0000-0003-4600-7331} \and Dominique Mercier\inst{1}\inst{2}\printfnsymbol{1} \and Andreas Dengel\inst{1}\inst{2}\orcidID{0000-0002-6100-8255} \and Sheraz Ahmed\inst{1}\orcidID{0000-0002-4239-6520}}
\authorrunning{S. A. Siddiqui et al.}
%
\institute{German Research Center for Artificial Intelligence (DFKI), 67663 Kaiserslautern, Germany \\ 
\email{first\_name.last\_name@dfki.de} \and
TU Kaiserslautern, 67663 Kaiserslautern, Germany}
\maketitle              
\begin{abstract}
With the rise in the employment of deep learning methods in safety-critical scenarios, interpretability is more essential than ever before. Although many different directions regarding interpretability have been explored for visual modalities, time-series data has been neglected with only a handful of methods tested due to their poor intelligibility. We approach the problem of interpretability in a novel way by proposing TSInsight where we attach an auto-encoder to the classifier with a sparsity-inducing norm on its output and fine-tune it based on the gradients from the classifier and a reconstruction penalty. TSInsight learns to preserve features that are important for prediction by the classifier and suppresses those that are irrelevant i.e. serves as a feature attribution method to boost interpretability. In contrast to most other attribution frameworks, TSInsight is capable of generating both instance-based and model-based explanations. We evaluated TSInsight along with 9 other commonly used attribution methods on 8 different time-series datasets to validate its efficacy. Evaluation results show that TSInsight naturally achieves output space contraction, therefore, is an effective tool for the interpretability of deep time-series models.
\end{abstract}

\section{Introduction}

Deep learning models have been at the forefront of technology in a range of different domains including image classification~\cite{NIPS2012_AlexNet}, object detection~\cite{girshick2015fast}, speech recognition~\cite{NIPS2010_Phone_Recognition}, text recognition~\cite{ocropus} and image captioning~\cite{karpathy2015deep}. 
These models are particularly effective in automatically discovering useful features. However, this automated feature extraction comes at the cost of lack of transparency of the system. 
Therefore, despite these advances, their employment in safety-critical domains like finance~\cite{mitTechReview2017}, self-driving cars~\cite{textualExplanationsForSelfDrivingCars} and medicine~\cite{mriDataPredictiveDifferenceAnalysis} is limited due to the lack of interpretability of the decision made by the network. 

Numerous efforts have been made for the interpretation of these black-box models. These efforts can be mainly classified into two separate directions. The first set of strategies focuses on making the network itself interpretable by trading off some performance. These strategies include Self-Explainable Neural Network (SENN)~\cite{senn2018} and Bayesian non-parametric regression models~\cite{guo2018explaining}. The second set of strategies focuses on explaining a pretrained model i.e. they try to infer the reason for a particular prediction. These attribution techniques include saliency map~\cite{yosinski15} and layer-wise relevance propagation~\cite{LRP_Bach_15}. However, all of these methods have been particularly developed and tested for visual modalities which are directly intelligible for humans. 
Transferring methodologies developed for visual modalities to time-series data is difficult due to the non-intuitive nature of time-series. Therefore, only a handful of methods have been focused on explaining time-series models in the past~\cite{timeseries-viz,tsviz}.



We approach the attribution problem in a novel way by attaching an auto-encoder on top of the classifier. The auto-encoder is fine-tuned based on the gradients from the classifier. Rather than optimizing the auto-encoder to reconstruct the whole input, we optimize the network to only reconstruct parts which are useful for the classifier i.e. are correlated or causal for the prediction. In order to achieve this, we introduce a sparsity inducing norm onto the output of the auto-encoder. In particular, the contributions of this paper are twofold:
\begin{itemize}
    \item A novel attribution method for time-series data which makes it much easier to interpret the decision of any deep learning model. The method also leverages dataset-level insights when explaining individual decisions in contrast to other attribution methods.
    \item Detailed analysis of the information captured by 11 different attribution techniques using suppression test on 8 different time-series datasets. This also includes analysis of the different out of the box properties achieved by TSInsight including generic applicability and contraction in the output space. 
\end{itemize}


\section{Related Work} \label{sec:related}

Since the resurgence of deep learning in 2012 after a deep network comprehensively outperformed its feature engineered counterparts~\cite{NIPS2012_AlexNet} on the ImageNet visual recognition challenge comprising of 1.2 million images~\cite{ILSVRC15}, deep learning has been integrated into a range of different applications to gain unprecedented levels of improvement. Significant efforts have been made in the past regarding the interpretability of deep models, specifically for image modality. These methods are mainly categorized into two different streams where the first stream is focused on explaining the decisions of a pretrained network while the second stream is directed towards making models more interpretable by trading off accuracy. 

The first stream for explainable systems which attempts to explain pretrained models using attribution techniques has been a major focus of research in the past years. The most common strategy is to visualize the filters of the deep model~\cite{zeiler13,simonyanVZ13,yosinski15,palacio2018deep,LRP_Bach_15}. This is very effective for visual modalities since images are directly intelligible for humans.~\cite{zeiler13} introduced deconvnet layer to understand the intermediate representations of the network. They not only visualized the network, but were also able to improve the network based on these visualizations to achieve state-of-the-art performance on ImageNet~\cite{ILSVRC15}.~\cite{simonyanVZ13} proposed a method to visualize class-specific saliency maps.~\cite{yosinski15} developed a visualization framework for image-based deep learning models. They tried to visualize the features that a particular filter was responding to by using regularized optimization. Instead of using first-order gradients, ~\cite{LRP_Bach_15} introduced a Layer-wise Relevance Propagation (LRP) framework which identified the relevant portions of the image by distributing the contribution to the incoming nodes.~\cite{smoothGrad} introduced the SmoothGrad method where they computed the mean gradients after adding small random noise sampled from a zero-mean Gaussian distribution to the original point.~\cite{integratedGradients} introduced the Integrated gradients method which works by computing the average gradient from the original point to the baseline input (zero-image in their case) at regular intervals.~\cite{guo2018explaining} used Bayesian non-parametric regression mixture model with multiple elastic nets to extract generalizable insights from the trained model. 
Recently, \cite{extremal_perturb} presented the extremal perturbation method where they solve an optimization problem to discover the minimum enclosing mask for an image that retains the network's predictive performance. 
Either these methods are not directly applicable to time-series data, or are inferior in terms of intelligibility for time-series data.

\cite{palacio2018deep} introduced yet another approach to understand a deep model by leveraging auto-encoders. 
After training both the classifier and the auto-encoder in isolation, they attached the auto-encoder to the head of the classifier and fine-tuned only the decoder freezing the parameters of the classifier and the encoder. This transforms the decoder to focus on features which are relevant for the network. Applying this method directly to time-series yields no interesting insights (Fig.~\ref{fig:auto_encoder_formulation}c) into the network's preference for input. Therefore, this method is strictly a special case of the TSInsight's formulation. 

In the second stream for explainable systems, ~\cite{senn2018} proposed Self-Explaining Neural Networks (SENN) where they learn two different networks. The first network is the concept encoder which encodes different concepts while the second network learns the weightings of these concepts. This transforms the system into a linear problem with a set of features making it easily interpretable for humans. SENN trade-offs accuracy in favor of interpretability.~\cite{textualExplanationsForSelfDrivingCars} attached a second network (video-to-text) to the classifier which was responsible for the production of natural language based explanation of the decisions taken by the network using the saliency information from the classifier. This framework relies on LSTM for the generation of the descriptions adding yet another level of opaqueness making it hard to decipher whether the error originated from the classification network or from the explanation generator.

\cite{timeseries-viz} made the first attempt to understand deep learning models for time-series analysis where they specifically focused on financial data. They computed the input saliency based on the first-order gradients of the network.~\cite{tsviz} proposed an influence computation framework which enabled exploration of the network at the filter level by computing the per filter saliency map and filter importance again based on first-order gradients. However, both methods lack in providing useful insights due to the noise inherent to first-order gradients. Another major limitation of saliency based methods is the sole use of local information. 
Therefore, TSInsight significantly supersedes in the identification of the important regions of the input using a combination of both local information for that particular example along with generalizable insights extracted from the entire dataset in order to reach a particular description.

Due to the use of auto-encoders, TSInsight is inherently related to sparse~\cite{ng2011sparse} and contractive auto-encoders~\cite{rifai2011contractive}. In sparse auto-encoders~\cite{ng2011sparse}, the sparsity is induced on the hidden representation by minimizing the KL-divergence between the average activations and a hyperparameter which defines the fraction of non-zero units. This KL-divergence is a necessity for sigmoid-based activation functions. However, in our case, the sparsity is induced directly on the output of the auto-encoder, which introduces a contraction on the input space of the classifier, and can directly be achieved by using Manhattan norm on the activations as we obtain real-valued outputs. Albeit sparsity being introduced in both cases, the sparsity in the case of sparse auto-encoders is not useful for interpretability. In the case of contractive auto-encoders~\cite{rifai2011contractive}, a contraction mapping is introduced by penalizing the Fobenius norm of the Jacobian of the encoder along with the reconstruction error. This makes the learned representation invariant to minor perturbations in the input. TSInsight on the other hand, induces a contraction on the input space for interpretability, thus, favoring sparsity inducing norm.


\section{Method} \label{sec:method}

\begin{wrapfigure}{r}{0.4\linewidth}
    \centering
    \vspace{-\intextsep}
    \includegraphics[width=1.0\linewidth]{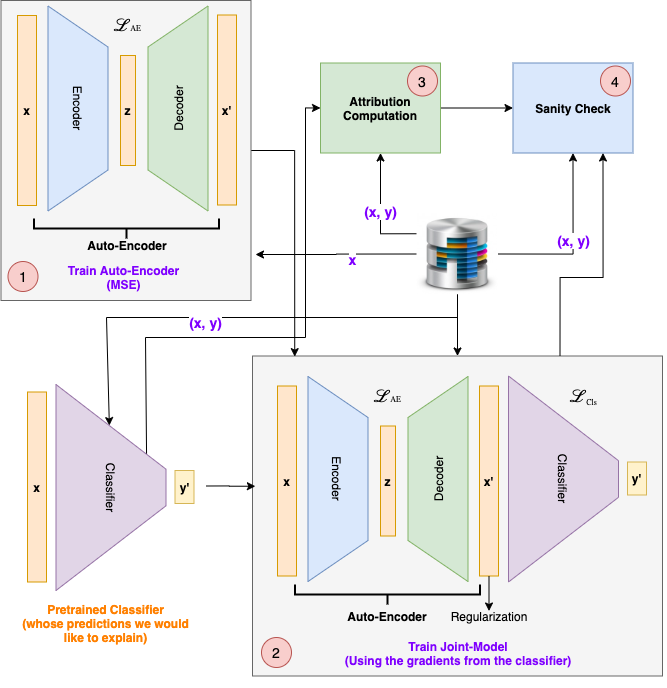}
    \caption{System Pipeline}
    \label{fig:sysOverview}
    \vspace{-\intextsep}
\end{wrapfigure}

The overview of our methodology is presented in Fig.~\ref{fig:sysOverview}. As the purpose of TSInsight is to explain the predictions of a pretrained model, we train a vanilla auto-encoder on the desired dataset as the first step. Once the auto-encoder is trained, we stack auto-encoder on top of the pretrained classifier to obtain a combined model.
We then only fine-tune the auto-encoder within the combined model using the gradients from the classifier using a specific loss function to highlight the causal/correlated points. We will first cover some basic background and then dive into the formulation of the problem presented by Palacio et al.~\cite{palacio2018deep}. We will then present the proposed formulation adapting the basic one for the interpretability of deep learning based time-series models.


\subsection{Pretrained Classifier}

A classifier ($\Phi: \mathcal{X} \mapsto \mathcal{Y}$) is a mapping from the input space $\mathcal{X}$ to the output space $\mathcal{Y}$. 
As the emphasis of TSInsight is interpretability, we assume the presence of a pretrained classifier whose predictions we are willing to explain. 
For this purpose, we trained a classifier using standard empirical risk minimization on the given dataset. The objective for the classifier training can be represented as:

\begin{equation}
\centering
    \mathcal{W}^{*} = \argmin{\mathcal{W}} \frac{1}{|\mathcal{X}|} \sum_{(\mathbf{x}, y) \in \mathcal{X} \times \mathcal{Y}} \mathcal{L}\big(\Phi(\mathbf{x}; \mathcal{W}^*), y\big) + \lambda \norm{\mathcal{W}}_2^2
    \label{eq:train}
\end{equation}

\noindent where $\Phi$ defines the mapping from the input space $\mathcal{X}$ to the output space $\mathcal{Y}$. 

\subsection{Auto-Encoder} 

    

An auto-encoder ($D \circ E: \mathcal{X} \mapsto \mathcal{X}$) is a neural network where the defined objective is to reconstruct the provided input by embedding it into an arbitrary feature space $\mathcal{F}$, therefore, is a mapping from the input space $\mathcal{X}$ to the input space itself $\mathcal{X}$ after passing it through the feature space $\mathcal{F}$. The auto-encoder is usually trained through mean-squared error as the loss function. The optimization problem for an auto-encoder can be represented as:

\begin{equation}
    (\mathcal{W}_{E}^{*}, \mathcal{W}_{D}^{*}) = \argmin{\mathcal{W}_{E}, \mathcal{W}_{D}} \frac{1}{|\mathcal{X}|} \sum_{\mathbf{x} \in \mathcal{X}} \norm{\mathbf{x} - D\big(E(\mathbf{x}; \mathcal{W}_{E}); \mathcal{W}_{D}\big)}_2^2 + \lambda \big(\norm{\mathcal{W}_{E}}_2^2 + \norm{\mathcal{W}_{D}}_2^2\big)
    \label{eq:autoEnc}
\end{equation}

\noindent where $E$ defines the encoder with parameters $\mathcal{W}_{E}$ while $D$ defines the decoder with parameters $\mathcal{W}_{D}$. Similar to the case of classifier, we train the auto-encoder using empirical risk minimization on a particular dataset. A sample reconstruction from the auto-encoder is visualized in Fig.~\ref{fig:auto_encoder_formulation}b for the forest cover dataset. It can be seen that the network did a reasonable job in the reconstruction of the input.

\subsection{Formulation by Palacio et al.~\cite{palacio2018deep}}

Palacio et al. (2018)~\cite{palacio2018deep} presented an approach for discovering the preference the network had for the input by attaching the auto-encoder on top of the classifier. The auto-encoder was fine-tuned using the gradients from the classifier.
The new optimization problem for fine-tuning the auto-encoder can be represented as:

\begin{multline}
    (\mathcal{W}_{E}^{'}, \mathcal{W}_{D}^{'}) = \argmin{\mathcal{W}_{E}^{*}, \mathcal{W}_{D}^{*}} \frac{1}{|\mathcal{X}|} \sum_{(\mathbf{x}, y) \in \mathcal{X} \times \mathcal{Y}} \mathcal{L}\bigg(\Phi\Big(D\big(E(\mathbf{x}; \mathcal{W}_{E}^{*}); \mathcal{W}_{D}^{*}\big); \mathcal{W}^*\Big), y\bigg) \\ + \lambda \big(\norm{\mathcal{W}_{E}^{*}}_2^2 + \norm{\mathcal{W}_{D}^{*}}_2^2\big)
    \label{eq:fineTune}
\end{multline}

\noindent where $\mathcal{W}_{E}^{*}$ and $\mathcal{W}_{D}^{*}$ are initialized from the auto-encoder weights obtained after solving the optimization problem specified in Eq.~\ref{eq:autoEnc} while $\mathcal{W}^{*}$ is obtained by solving the optimization problem specified in Eq.~\ref{eq:train}. 
This formulation is slightly different from the one proposed by Palacio et al. (2018) where they only fine-tuned the decoder part of the auto-encoder, while we update both the encoder as well as the decoder as it is a much natural formulation as compared to only fine-tuning the decoder. This complete fine-tuning is significantly more important once we move towards advanced formulations since we would like the network to also adapt the encoding in order to better focus on important features. Fine-tuning only the decoder will only change the output without the network learning to compress the signal itself.

\subsection{TSInsight: The Proposed Formulation} 

\begin{figure}[t!p]
    \centering
    \includegraphics[width=0.9\linewidth]{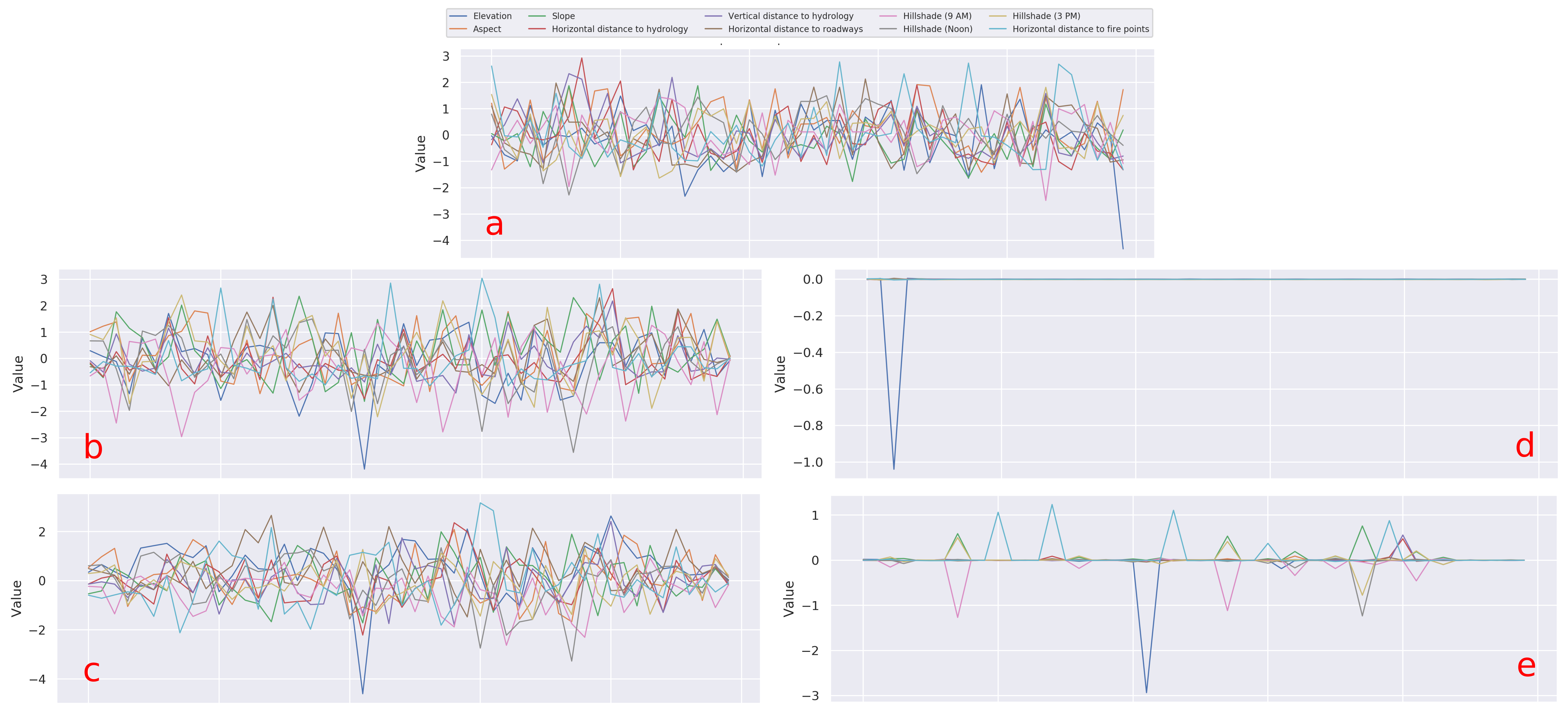}
    \caption{Comparison of different auto-encoder outputs (a) Original input, (b) Reconstruction from the vanilla auto-encoder, (c) Palacio et al.~\cite{palacio2018deep} (d) Auto-encoder finetuned with sparsity, and (e) TSInsight.}
    \label{fig:auto_encoder_formulation}
\end{figure}

In contrast to the findings of Palacio et al. (2018)~\cite{palacio2018deep} for the image domain, directly optimizing the objective defined in Eq.~\ref{eq:fineTune} for time-series yields no interesting insights into the input preferred by the network. This effect is amplified with the increase in the dataset complexity. 
Fig.~\ref{fig:auto_encoder_formulation}c presents an example from the forest cover dataset. It is evident from the figure that the resulting reconstruction from the fine-tuned auto-encoder provides no useful insights regarding the causality of points for a particular prediction. Therefore, instead of optimizing this raw objective, we modify the objective by adding the sparsity-inducing norm on the output of the auto-encoder. Inducing sparsity on the auto-encoder's output forces the network to only reproduce relevant regions of the input to the classifier since the auto-encoder is optimized using the gradients from the classifier. 
However, just optimizing for sparsity introduces misalignment between the reconstruction and the input as visualized in Fig.~\ref{fig:auto_encoder_formulation}d. In order to ensure alignment between the two sequences, we additionally introduce the reconstruction loss into the final objective.
Therefore, the proposed TSInsight optimization objective can be written as:

\begin{multline}
    (\mathcal{W}_{E}^{'}, \mathcal{W}_{D}^{'}) = \argmin{\mathcal{W}_{E}^{*}, \mathcal{W}_{D}^{*}} \frac{1}{|\mathcal{X}|}  \sum_{(\mathbf{x}, y) \in \mathcal{X} \times \mathcal{Y}} \Bigg[ \mathcal{L}\bigg(\Phi\Big(D\big(E(\mathbf{x}; \mathcal{W}_{E}^{*}); \mathcal{W}_{D}^{*}\big); \mathcal{W}^{*}\Big), y\bigg) + \\ \gamma \Big(\norm{\mathbf{x} - D\big(E(\mathbf{x}; \mathcal{W}_{E}^{*}); \mathcal{W}_{D}^{*}\big)}_2^2 \Big) + \beta \Big(\norm{D\big(E(\mathbf{x}; \mathcal{W}_{E}^{*}); \mathcal{W}_{D}^{*}\big)}_1 \Big) \Bigg] \\ + \lambda \big(\norm{\mathcal{W}_{E}^{*}}_2^2 + \norm{\mathcal{W}_{D}^{*}}_2^2 \big)
    \label{eq:regReconsFineTune}
\end{multline}

\noindent where $\mathcal{L}$ represents the classification loss function which is cross-entropy in our case, $\Phi$ denotes the classifier with pretrained weights $\mathcal{W}^*$, while $E$ and $D$ denotes the encoder and decoder respectively with corresponding pretrained weights $\mathcal{W}_{E}^*$ and $\mathcal{W}_{D}^*$. We introduce two new hyperparameters, $\gamma$ and $\beta$. $\gamma$ controls the auto-encoder's focus on reconstruction of the input. $\beta$ on the other hand, controls the sparsity enforced on the output of the auto-encoder. 
After training the auto-encoder with the TSInsight objective function, the output is both sparse as well as aligned with the input as evident from Fig.~\ref{fig:auto_encoder_formulation}e.

The hyperparameters play an essential role for TSInsight to provide useful insights into the model's behavior. Performing grid search to determine this value is not possible as large values of $\beta$ results in models which are more interpretable but inferior in terms of performance, therefore, presenting a trade-off between performance and interpretability which is difficult to quantify. Although we found manual tuning of hyperparameters to be superior, we also investigated the employment of feature importance measures~\cite{tsviz,mfi} for the automated selection of these hyperparameters ($\beta$ and $\gamma$). The simplest candidate for this importance measure is saliency. This can be written as:

\begin{equation*} \label{eq:sal}
	I(\mathbf{x}) = \frac{\partial a^{L}} {\partial \mathbf{x}}
\end{equation*}

\noindent where $L$ denotes the number of layers in the classifier and $a^{L}$ denotes the activations of the last layer in the classifier. 
This saliency-based importance computation is only based on the classifier.
Once the corresponding importance values are computed, they are scaled in the range of [0, 1] to serve as the corresponding reconstruction weight i.e. $\gamma$. The inverted importance values then serve as the corresponding sparsity weight i.e. $\beta$.

\begin{equation*} \label{eq:minMaxSc}
	I(\mathbf{x}) = \frac{I(\mathbf{x}) - \min\limits_{j} I(\mathbf{x})_{j}} {\max\limits_{j} I(\mathbf{x})_{j} - \min\limits_{j} I(\mathbf{x})_{j}}
\end{equation*}

\begin{equation*} \label{eq:inversion}
	\gamma^{*}(\mathbf{x}) = I(\mathbf{x}) \qquad \& \qquad \beta^{*}(\mathbf{x}) = 1.0 - I(\mathbf{x})
\end{equation*}

\noindent Therefore, the final term imposing sparsity on the classifier can be written as:


\begin{multline*}
	\gamma \Big(\norm{\mathbf{x} - D\big(E(\mathbf{x}; \mathcal{W}_{E}^{*}); \mathcal{W}_{D}^{*}\big)}_2^2 \Big) + \beta \Big(\norm{D\big(E(\mathbf{x}; \mathcal{W}_{E}^{*}); \mathcal{W}_{D}^{*}\big)}_1 \Big) \Rightarrow \\ C \times \norm{D\big(E(\mathbf{x}; \mathcal{W}_{E}^{*}); \mathcal{W}_{D}^{*}\big) \odot \beta^{*}(\mathbf{x})}_1 + \norm{\Big(\mathbf{x} - D\big(E(\mathbf{x}; \mathcal{W}_{E}^{*}); \mathcal{W}_{D}^{*}\big)\Big) \odot \gamma^{*}(\mathbf{x})}_2^2
\end{multline*}

\noindent In contrast to the instance-based value of $\beta$, we used the average saliency value in our experiments.
This ensures that the activations are not sufficiently penalized so as to significantly impact the performance of the classifier.
Due to the low relative magnitude of the sparsity term, we scaled it by a constant factor $C$ (we used $C=10$ in our experiments).

\section{Experimental Setup} \label{sec:exp_setup}

This section will cover the evaluation setup that we used to establish the utility of TSInsight in comparison to other commonly used attribution techniques. We will first define the evaluation metric we used to compare different attribution techniques. Then we will discuss the 8 different datasets that we used in our experimental study followed by the 11 different attribution techniques that we compared.

\subsection{Evaluation Metric} \label{sec:suppression_test}

A commonly used metric to compare model attributions in visual modalities is via the pointing-game or suppression test~\cite{extremal_perturb}. Since the pointing game is not directly applicable to time-series data, we compare TSInsight with other attribution techniques using the suppression test.
Suppression test attempts to quantify the quality of the attribution by just preserving parts of the input that are considered to be important by the method. This suppressed input is then passed onto the classifier. If the selected points are indeed causal/correlated to the prediction generated by the classifier, no evident effect on the prediction should be observed. On the other hand, if the points highlighted by the attribution technique are not the most important ones for prediction, the network's prediction will change. It is important to note that unless there is a high amount of sparsity present in the signal, suppressing the signal itself will result in a loss of accuracy for the classifier since there is a slight mismatch for the classifier for the inputs seen during training. We compared TSInsight with a range of different saliency methods. 

\subsection{Datasets} \label{sec:dataset}

We employed 8 different time-series dataset in our study. The summary of these datasets is available in Table~\ref{tab:datasets}. We will now cover each of these datasets in detail.

\begin{table*}[t]
\scriptsize
    \centering
    \caption{Dataset details}
    \begin{tabular}{c c c c c c c}
        \toprule
         \textbf{Dataset} & \textbf{Train} & \textbf{Validation} & \textbf{Test} & \textbf{Seq. Length} & \textbf{Input Ch.} & \textbf{\# Classes} \\
         \midrule
         Synthetic Anomaly Detection & 45000 & 5000 & 10000 & 50 & 3 & 2 \\
         Electric Devices & 6244 & 2682 & 7711 & 50 & 3 & 7 \\
         Character Trajectories & 1383 & 606 & 869 & 206 & 3 & 20 \\
         FordA & 2520 & 1081 & 1320 & 500 & 1 & 2 \\
         Forest Cover & 107110 & 45906 & 65580 & 50 & 10 & 2 \\
         ECG Thorax & 1244 & 556 & 1965 & 750 & 1 & 42 \\
         WESAD & 5929 & 846 & 1697 & 700 & 8 & 3 \\
         UWave Gesture & 624 & 272 & 3582 & 946 & 1 & 8 \\
         \bottomrule
    \end{tabular}
    \label{tab:datasets}
\end{table*}

\noindent \textbf{Synthetic Anomaly Detection Dataset:}
The synthetic anomaly detection dataset~\cite{tsviz} is a synthetic dataset comprising of three different channels referring to the pressure, temperature and torque values of a machine running in a production setting where the task is to detect anomalies. The dataset only contains point-anomalies. If a point-anomaly is present in a sequence, the whole sequence is marked as anomalous. Anomalies were intentionally never introduced on the pressure signal in order to identify the treatment of the network to that particular channel. 

\noindent \textbf{Electric Devices Dataset:}
The electric devices dataset~\cite{electricDevices} is a small subset of the data collected as part of the UK government's sponsored study, \textit{Powering the Nation}. The aim of this study was to reduce UK's carbon footprint. The electric devices dataset is comprised of data from 251 households, sampled in two-minute intervals over a month. 

\noindent \textbf{Character Trajectories Dataset:}
The character trajectories dataset\footnote{\url{https://archive.ics.uci.edu/ml/datasets/Character+Trajectories}} contains hand-written characters using a Wacom tablet. Only three dimensions are kept for the final dataset which includes x, y and pen-tip force. 
The sampling rate was set to be 200 Hz. The data was numerically differentiated and Gaussian smoothen with $\sigma = 2$. The task is to classify the characters into 20 different classes.

\noindent \textbf{FordA Dataset:}
The FordA dataset\footnote{\url{http://www.timeseriesclassification.com/description.php?Dataset=FordA}} was originally used for a competition organized by IEEE in the IEEE World Congress on Computational Intelligence (2008). It is a binary classification problem where the task is to identify whether a certain symptom exists in the automotive subsystem. FordA dataset was collected with minimal noise contamination in typical operating conditions. 

\noindent \textbf{Forest Cover Dataset:}
The forest cover dataset~\cite{forestCoverAnomaly} has been adapted from the UCI repository for the classification of forest cover type from cartographic variables. The dataset has been transformed into an anomaly detection dataset by selecting only 10 quantitative attributes out of a total of 54. Instances from the second class were considered to be normal while instances from the fourth class were considered to be anomalous. The ratio of the anomalies to normal data points is 0.9\%. Since only two classes were considered, the rest of them were discarded. 

\noindent \textbf{WESAD Dataset:}
WESAD dataset~\cite{boschDataset} is a classification dataset introduced by Bosch for person's affective state classification with three different classes, namely, neutral, amusement and stress. 

\noindent \textbf{ECG Thorax Dataset:}
The non-invasive fetal ECG Thorax dataset\footnote{\url{http://www.timeseriesclassification.com/description.php?Dataset=NonInvasiveFetalECGThorax1}} is a classification dataset comprising of 42 classes. 

\noindent \textbf{UWave Gesture Dataset:}
The wave gesture dataset~\cite{uWave} contains accelerometer data where the task is to recognize 8 different gestures. 

\subsection{Attribution Techniques} \label{sec:attribution_techniques}

We compared TSInsight against a range a commonly employed attribution techniques. Each attribution method provided us with an estimate of the features' importance which we used to suppress the signal. 
In all of the cases, we used the absolute magnitude of the corresponding feature attribution method to preserve the most-important input features.
Two methods i.e. $\epsilon-LRP$ and DeepLift were shown to be similar to $input \odot gradient$~\cite{adebayo2018sanity}, therefore, we compare only against $input \odot gradient$. We don't compute class-specific saliency, but instead, compute the saliency w.r.t. all the output classes. For all the methods computing class specific activations maps e.g. GradCAM, guided GradCAM, and occlusion sensitivity, we used the class with the maximum predicted score as our target.
The description of the 11 different attribution techniques evaluated in this study is provided below:

\noindent \textbf{None:} None refers to the absence of any importance measure. Therefore, in this case, the complete input is passed on to the classifier without any suppression for comparison.

\noindent \textbf{Random:} Random points from the input are suppressed in this case.

\noindent \textbf{Input Magnitude:} We treat the absolute magnitude of the input to be a proxy for the features' importance.

\noindent \textbf{Occlusion sensitivity:} We iterate over different input channels and positions and mask the corresponding input features with a filter size of 3 and compute the difference in the confidence score of the predicted class (i.e. the class with the maximum score on the original input). We treat this sensitivity score as the features' importance. This is a brute-force measure of feature importance and employed commonly in prior literature as served as a strong baseline in our experiments~\cite{zeiler13}. A major limitation of occlusion sensitivity is its execution speed since it requires iterating over the complete input running inference numerous times.

\noindent \textbf{TSInsight:} We treat the absolute magnitude of the output from the auto-encoder of TSInsight as features' importance.

\noindent \textbf{Palacio et al.:} Similar to TSInsight, we use the absolute magnitude of the auto-encoder's output as the features' importance~\cite{palacio2018deep}.

\noindent \textbf{Gradient:} We use the absolute value of the raw gradient of the classifier w.r.t. to all of the classes as the features' importance~\cite{tsviz,timeseries-viz}.

\noindent \textbf{Gradient} $\odot$ \textbf{Input:} We compute the Hadamard (element-wise) product between the gradient and the input, and use its absolute magnitude as the features' importance~\cite{integratedGradients}.

\noindent \textbf{Integrated Gradients:} We use absolute value of the integrated gradient with 100 discrete steps between the input and the baseline (which was zero in our case) as the features' importance~\cite{integratedGradients}.

\noindent \textbf{SmoothGrad:} We use the absolute value of the smoothened gradient computed by using 100 different random noise vector sampled from a Gaussian distribution with zero mean, and a variance of $2 / (max_j \mathbf{x_j} - min_j \mathbf{x_j})$ where $\mathbf{x}$ was the input as the features' importance measure~\cite{smoothGrad}.

\noindent \textbf{Guided Backpropagation:} We use the absolute value of the gradient provided by guided backpropagation~\cite{guidedBackpropSpringenberg}. In this case, all the ReLU layers were replaced with guided ReLU layers which masks negative gradients, hence filtering out negative influences for a particular class to improve visualization.

\noindent \textbf{GradCAM:} We use the absolute value of Gradient-based Class Activation Map (GradCAM)~\cite{gradcam} as our feature importance measure. GradCAM computes the importance of the different filters present in the input in order to come up with a metric to score the overall output. Since GradCAM visualizes a class activation map, we used the predicted class as the target for visualization.

\noindent \textbf{Guided GradCAM:} Guided GradCAM~\cite{gradcam} is a guided variant of GradCAM which performs a Hadamard product (pointwise) of the signal from guided backpropagation and GradCAM to obtain guided GradCAM. We again use the absolute value of the guided GradCAM output as importance measure.

\section{Results} \label{sec:results}

\begin{table}[t!p]
    \tiny
    \centering
    \caption{Results for the different datasets in terms of accuracy for both the classifier as well as TSInsight.}
    \begin{tabular}{c c c c c}
        \toprule
         \textbf{Dataset} & \textbf{Model} & $\mathbf{\gamma}$ & $\mathbf{\beta}$ & \textbf{Accuracy} \\
         \midrule
         Synthetic Anomaly & Raw classifier & - & - & 98.01 \% \\
         Detection & TSInsight & 1.0 & 0.001 & 98.13 \% \\
         \midrule
         WESAD & Raw classifier & - & - & 99.94 \% \\
         & TSInsight & 2.0 & 0.00001 & 99.76 \% \\
         \midrule
         Character Trajectories & Raw classifier & - & - & 97.01 \% \\
         & TSInsight & 0.25 & 0.0001 & 97.24 \% \\
         \midrule
         FordA & Raw classifier & - & - & 91.74 \% \\
         & TSInsight & 2.0 & 0.0001 & 93.26 \% \\
         \midrule
         Forest Cover & Raw classifier & - & - & 95.79 \% \\
         & TSInsight & 4.0 & 0.0001 & 96.26 \% \\
         \midrule
         Electric Devices & Raw classifier & - & - & 65.14 \% \\
         & TSInsight & 4.0 & 0.0001 & 65.74 \% \\
         \midrule
         ECG Thorax & Raw classifier & - & - & 86.01 \% \\
         & TSInsight & 0.1 & 0.0001 & 84.07 \% \\
         \midrule
         UWave Gesture & Raw classifier & - & - & 91.76 \% \\
         & TSInsight & 4.0 & 0.0005 & 92.29 \% \\
         \bottomrule
    \end{tabular}
    \label{tab:results}
\end{table}

The results we obtained with the proposed formulation were highly intelligible for the datasets we employed in this study. TSInsight produced a sparse representation of the input focusing only on the salient regions. In addition to interpretability, with a careful tuning of the hyperparameters, TSInsight outperformed the pretrained classifier in terms of accuracy for most of the cases which is evident from Table~\ref{tab:results}. However, it is important to note that TSInsight is not designed for the purpose of performance, but rather for interpretability. Therefore, we expect that the performance will drop in many cases depending on the amount of sparsity enforced.

\begin{figure*}[t!p]
    \centering
    \includegraphics[width=\linewidth]{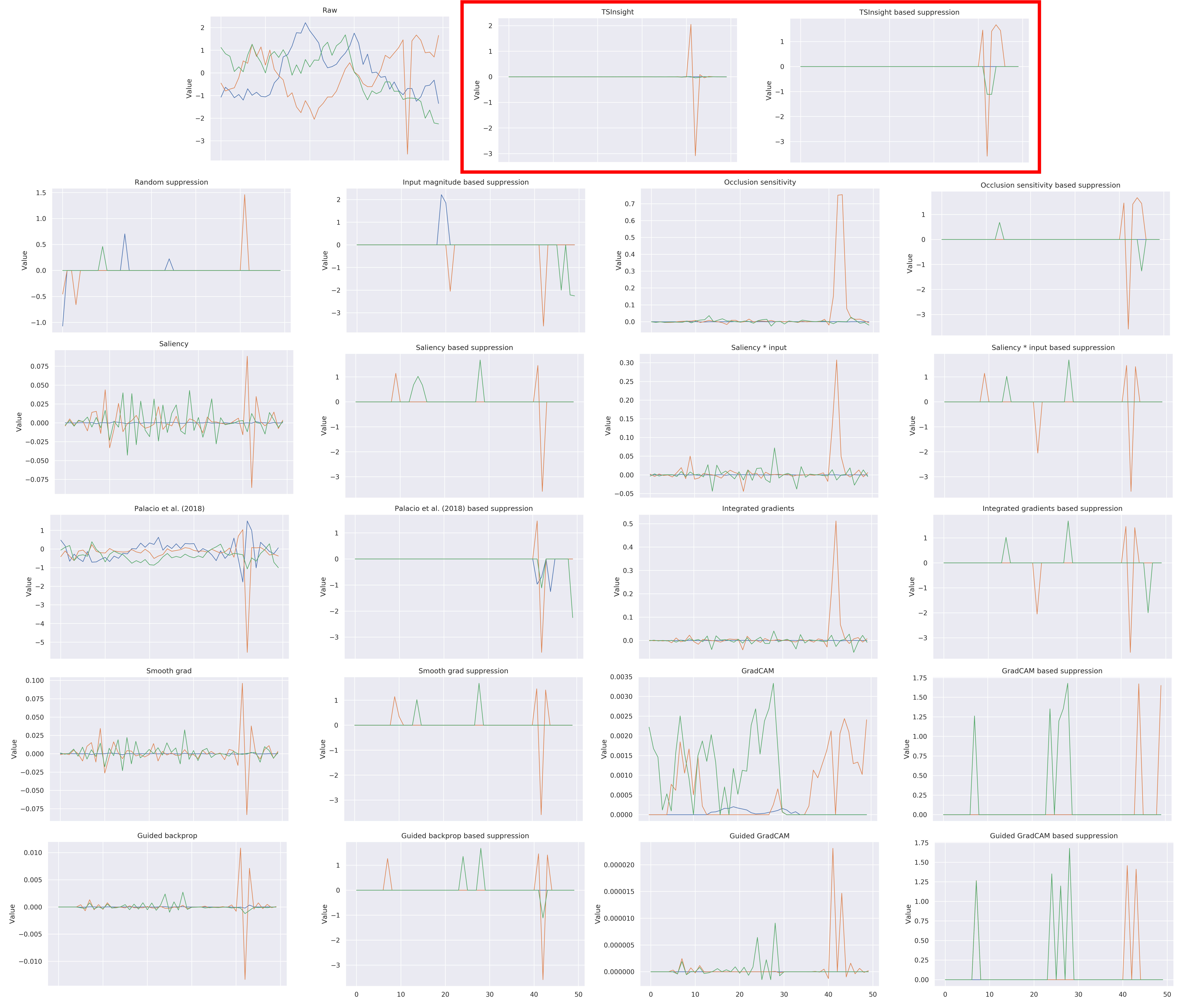}
    \caption{Output from different attribution methods as well as the input after suppressing all the points except the top 5\% highlighted by the corresponding attribution method on an anomalous example from the synthetic anomaly detection dataset (best viewed digitally).}
    \label{fig:sanityCheck}
\end{figure*}

In order to qualitatively assess the attribution provided by TSInsight, we visualize an anomalous example from the synthetic anomaly detection dataset in Fig.~\ref{fig:sanityCheck} along with the attributions from all the commonly employed attribution techniques (listed in Section~\ref{sec:attribution_techniques}). 
Since there were only a few relevant discriminative points in the case of forest cover and synthetic anomaly detection datasets, TSInsight suppressed most of the input making the decision directly interpretable. 

\begin{figure}[t!p]
    \centering
    \begin{subfigure}[b]{0.24\linewidth}
        \includegraphics[width=\linewidth]{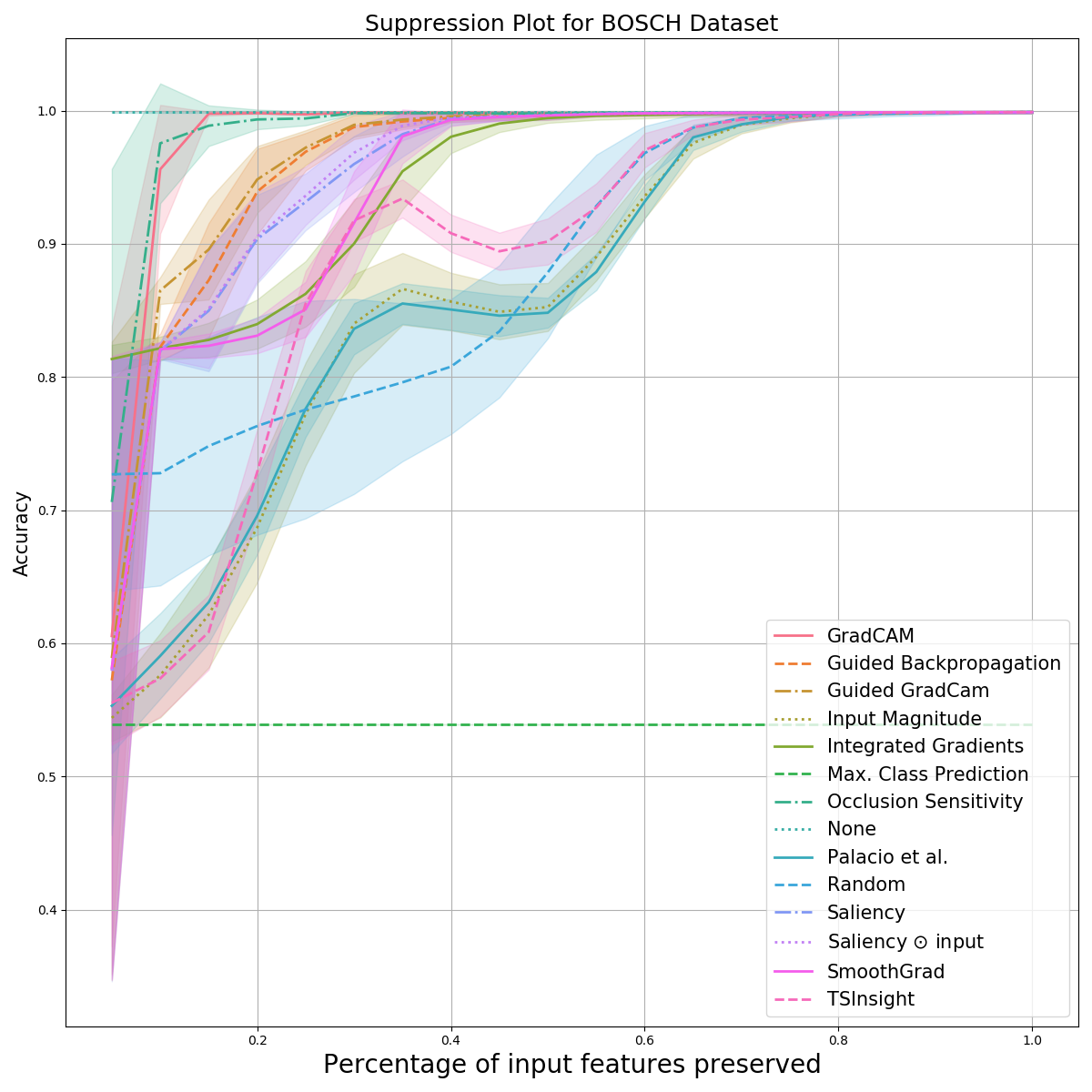}
        \caption{WESAD}
        \label{fig:suppresion_bosch}
    \end{subfigure}
    \begin{subfigure}[b]{0.24\linewidth}
    \includegraphics[width=\linewidth]{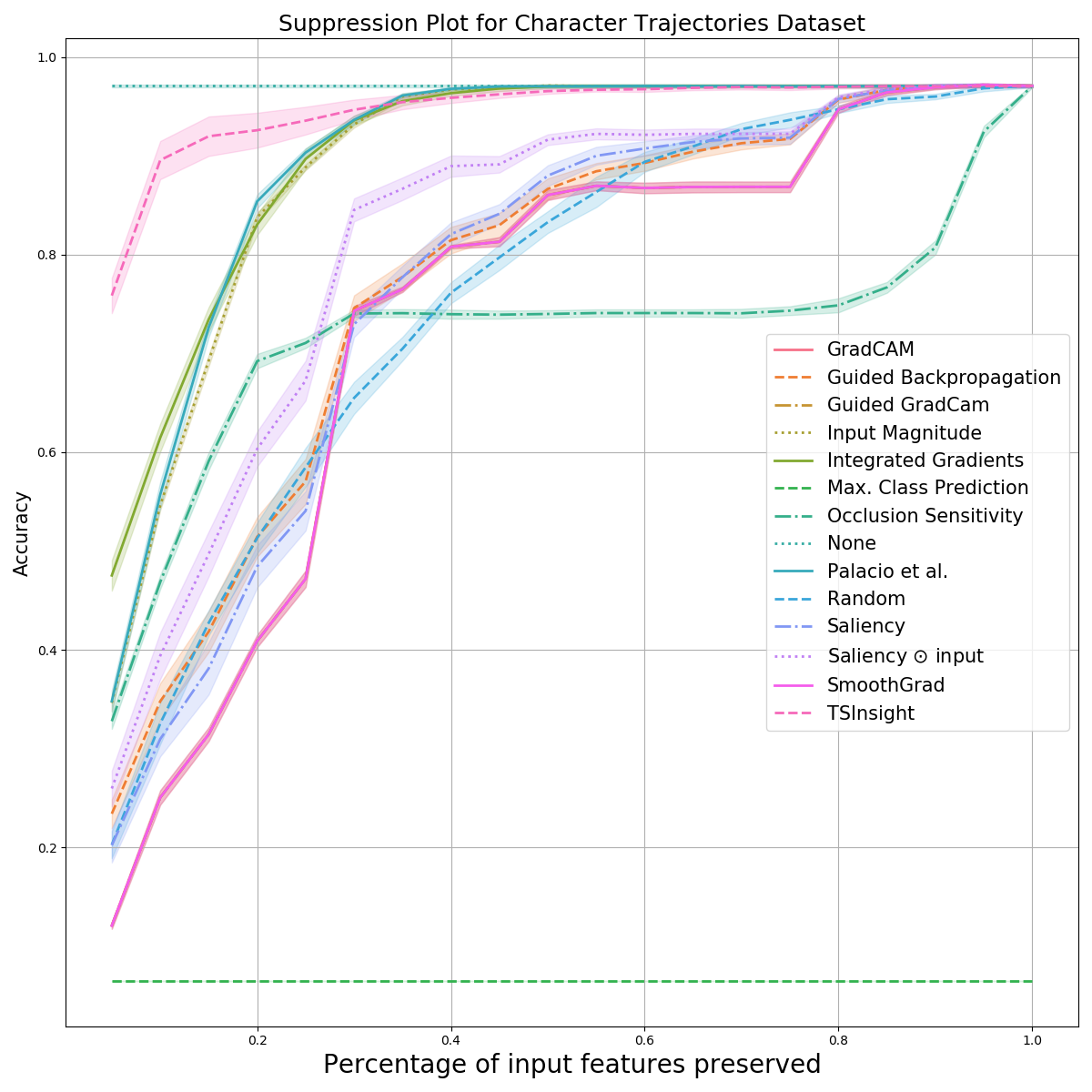}
        \caption{Character Traj.}
        \label{fig:suppression_character}
    \end{subfigure}
    \begin{subfigure}[b]{0.24\linewidth}
        \includegraphics[width=\linewidth]{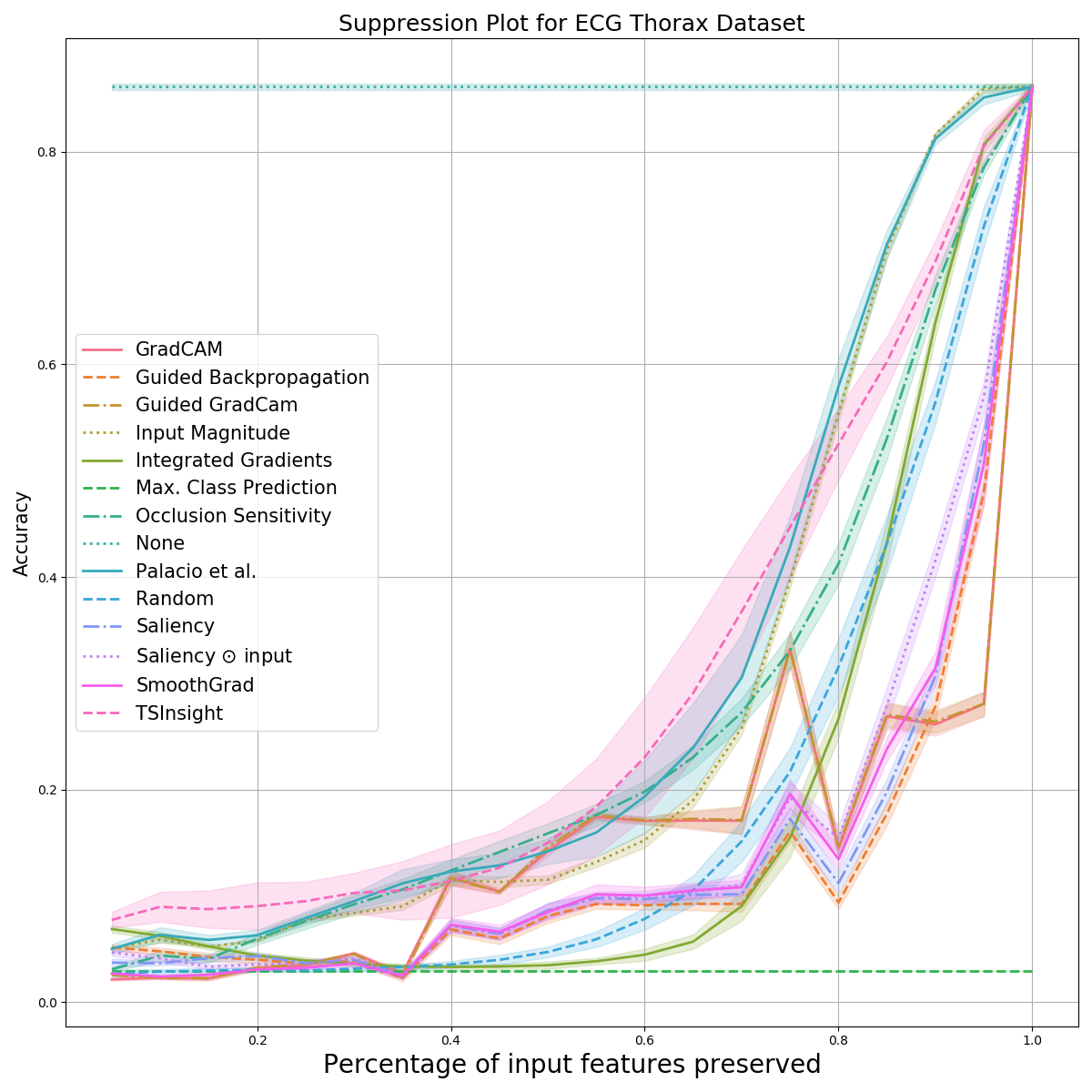}
        \caption{ECG Thorax}
        \label{fig:suppression_ecg}
    \end{subfigure}
    \begin{subfigure}[b]{0.24\linewidth}
        \includegraphics[width=\linewidth]{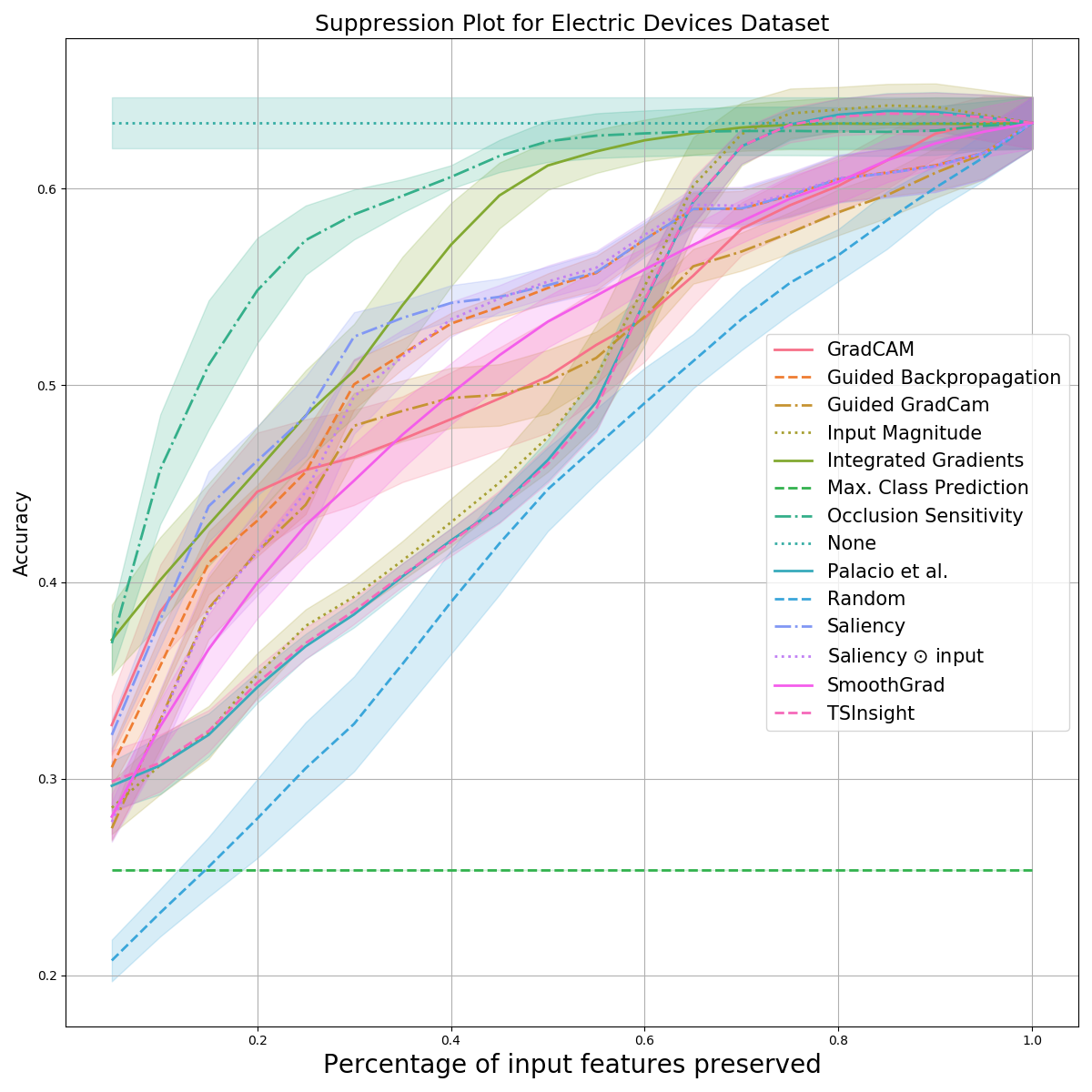}
        \caption{Electric Devices}
        \label{fig:suppression_electric}
    \end{subfigure}
    
    \begin{subfigure}[b]{0.24\linewidth}
        \includegraphics[width=\linewidth]{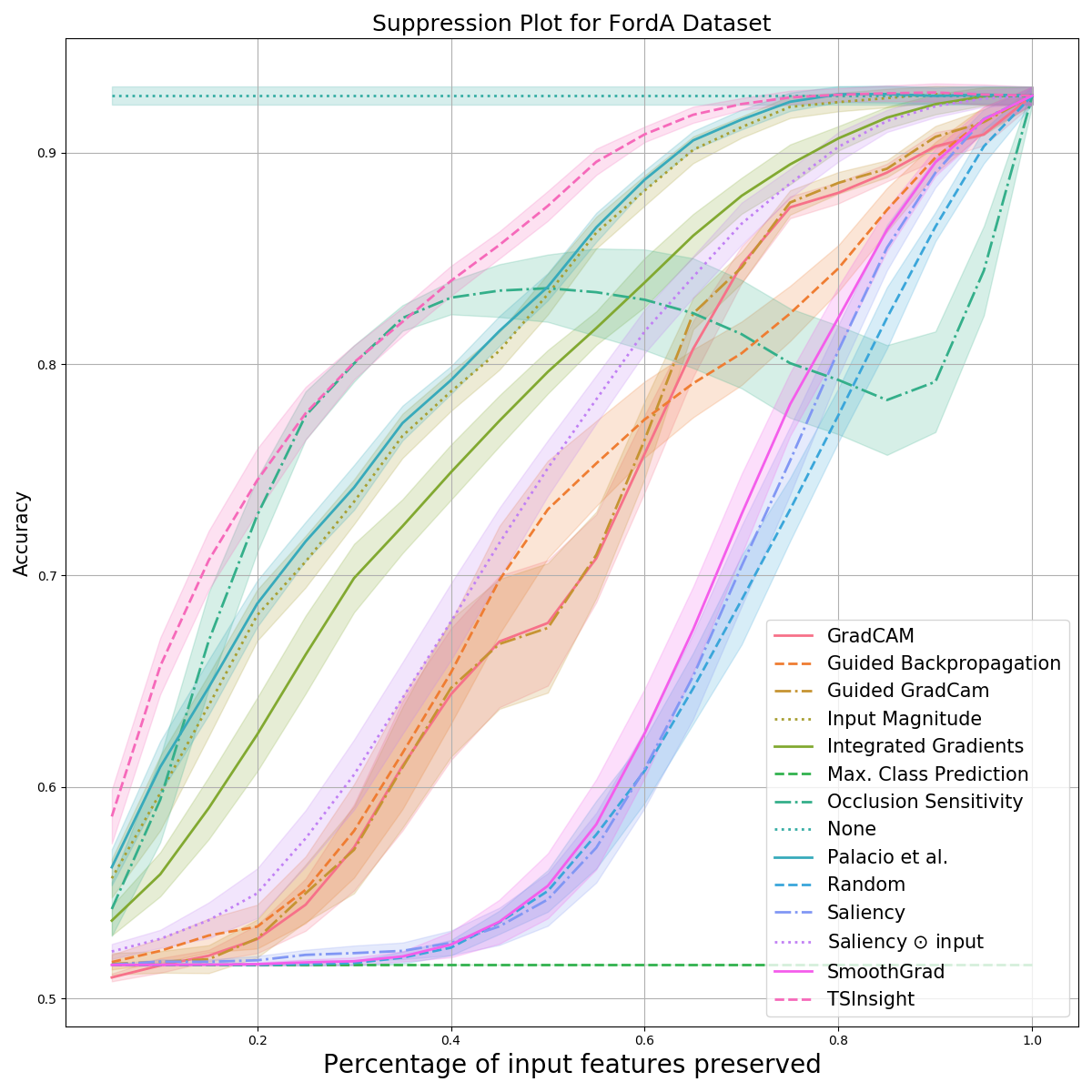}
        \caption{FordA}
        \label{fig:suppresion_fordA}
    \end{subfigure}
    \begin{subfigure}[b]{0.24\linewidth}
        \includegraphics[width=\linewidth]{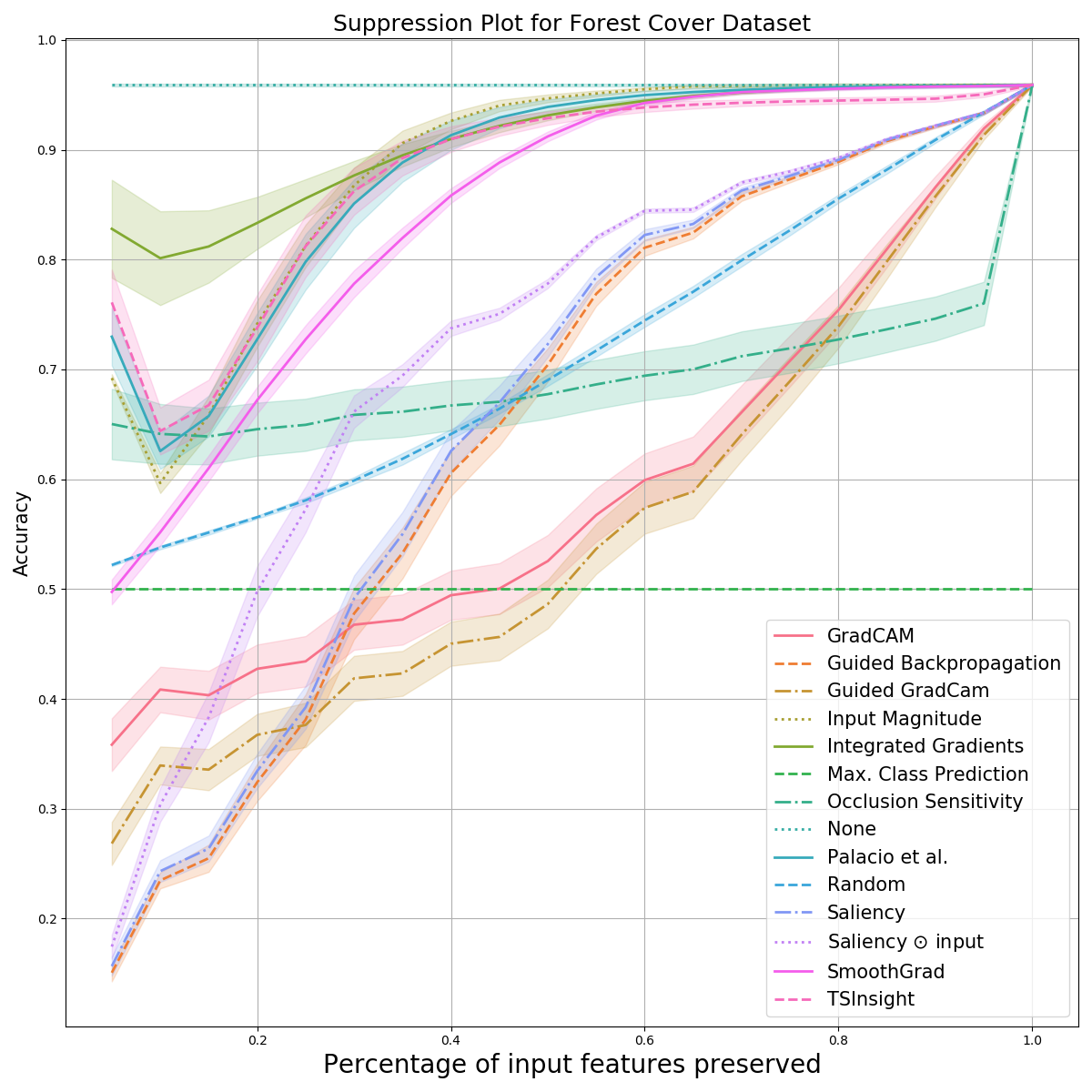}
        \caption{Forest Cover}
        \label{fig:suppression_forest}
    \end{subfigure}
    \begin{subfigure}[b]{0.24\linewidth}
        \includegraphics[width=\linewidth]{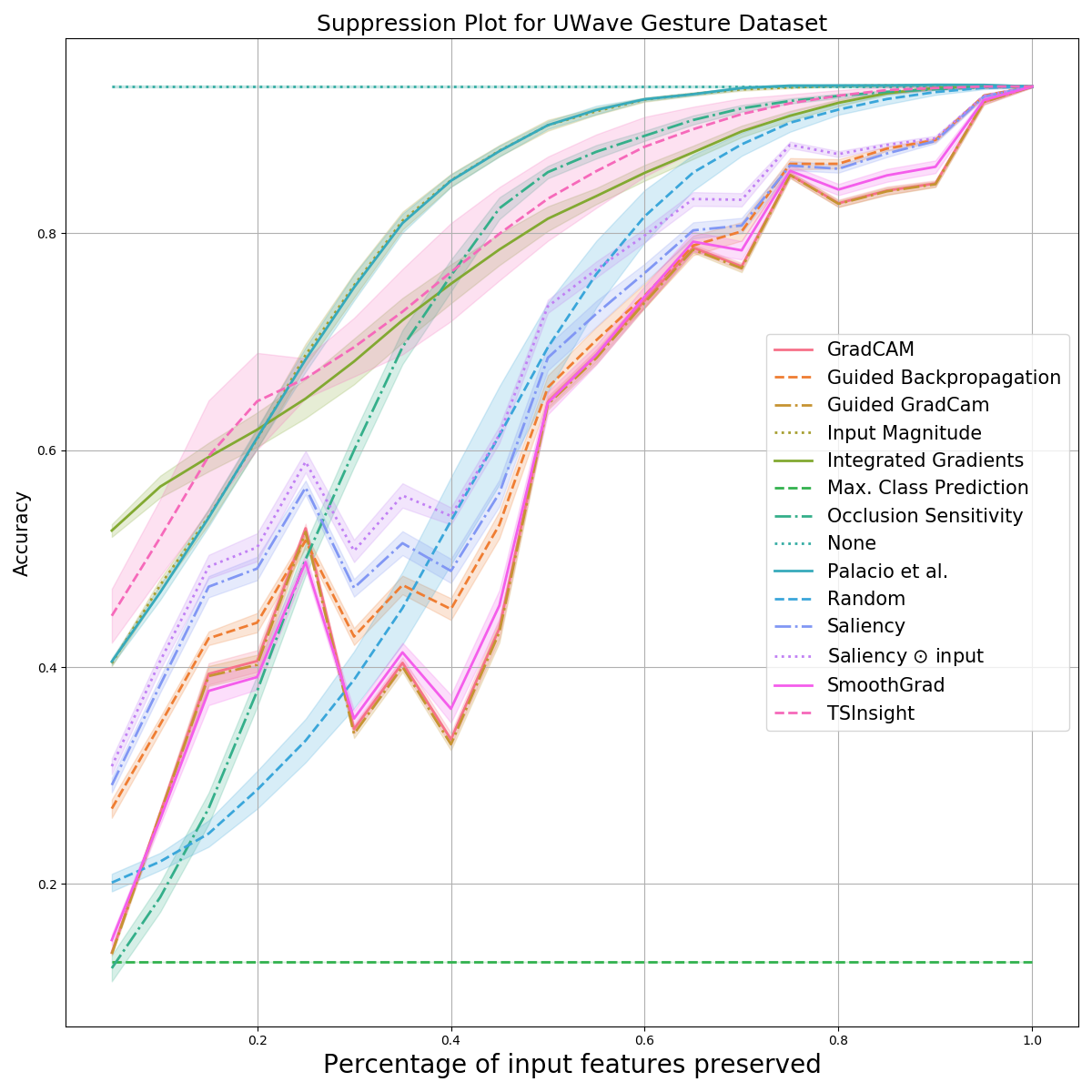}
        \caption{UWave Gesture}
        \label{fig:suppression_gesture}
    \end{subfigure}
    \begin{subfigure}[b]{0.24\linewidth}
        \includegraphics[width=\linewidth]{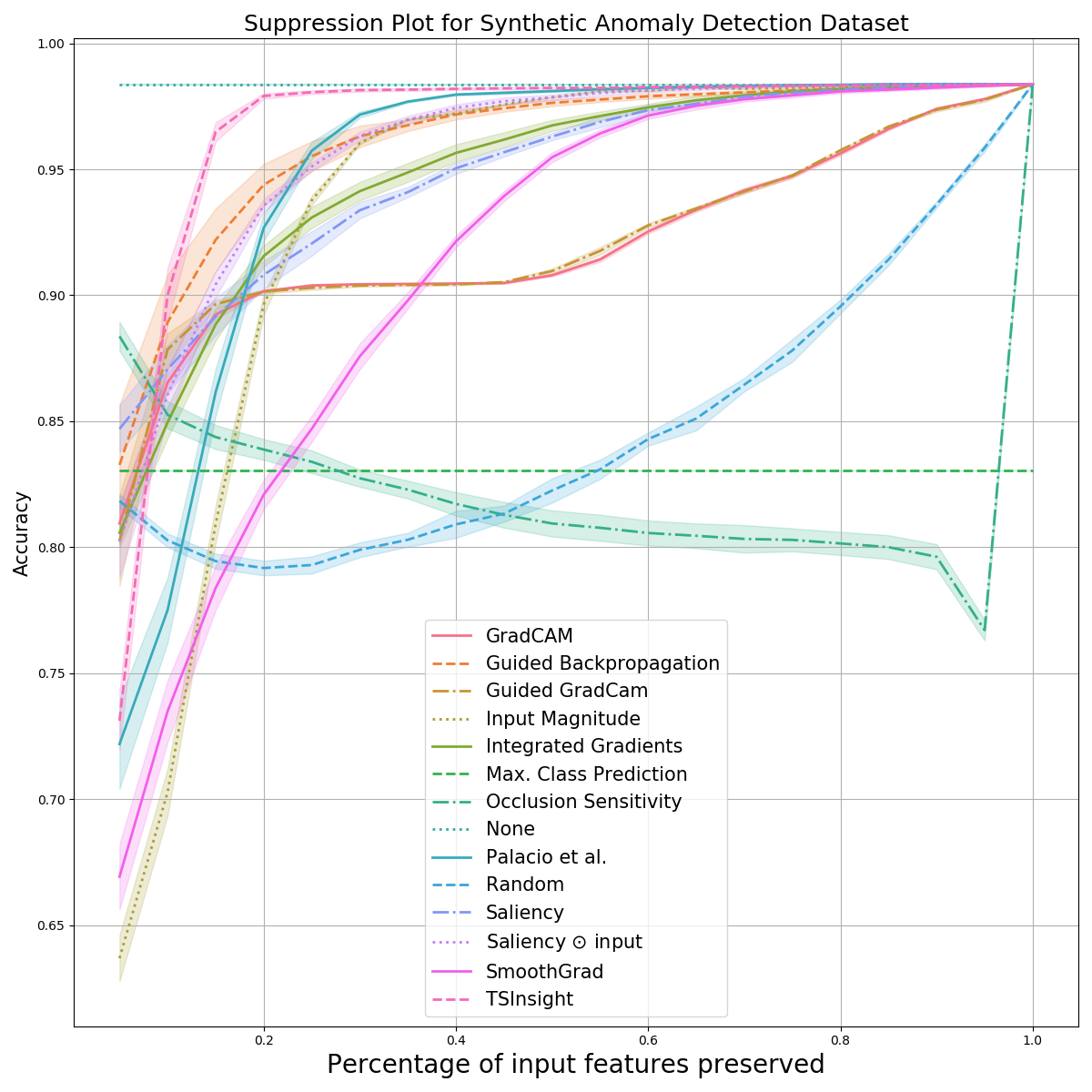}
        \caption{Anomaly}
        \label{fig:suppression_anomaly}
    \end{subfigure}
    
    \caption{Suppression results against a large number of baseline methods computed using 5 random runs (best viewed digitally).}
    \label{fig:suppression_plots}
\end{figure}

As described in Section~\ref{sec:suppression_test}, we compare the performance of different attribution techniques using the input suppression test. The results with different amount of suppression are visualized in Fig.~\ref{fig:suppression_plots} which are computed based on 5 random runs. Since the datasets were picked to maximize diversity in terms of the features, there is no perfect method which can perfectly generalize to all the datasets.
The different attribution techniques along with the corresponding suppressed input is visualized in Fig.~\ref{fig:sanityCheck} for the synthetic anomaly detection datasets. 
TSInsight produced the most plausible looking explanations along with being the most competitive saliency estimator on average in comparison to all other attribution techniques. 
Alongside the numbers, TSInsight was also able to produce the most plausible explanations.

\subsection{Properties of TSInsight} \label{sec:properties}

We will now discuss some of the interesting properties that TSInsight achieves out-of-the-box which includes output space contraction, its generic applicability and model-based (global) explanations. 
Since TSInsight induces a contraction in the input space, this also results in slight gains in terms of adversarial robustness. However, these gains are not consistent over many datasets and strong adversaries, therefore, omitted for clarity here. In depth evaluation of adversarial robustness of TSInsight can be an interesting future direction.

\subsubsection{Model-based vs Instance-based Explanations}

Since TSInsight poses the attribution problem itself as an optimization objective, the data based on which this optimization problem is solved defines the explanation scope. If the optimization problem is solved for the complete dataset, this tunes the auto-encoder to be a generic feature extractor, enabling extraction of model/dataset-level insights using the attribution. In contrary, if the optimization problem is solved for a particular input, the auto-encoder discovers an instance's attribution. This is contrary to most other attribution techniques which are only instance specific. 

\subsubsection{Auto-Encoder's Jacobian Spectrum Analysis}

\begin{wrapfigure}{r}{0.5\linewidth}
    \centering
    \vspace{-\intextsep}
    \includegraphics[width=1.0\linewidth]{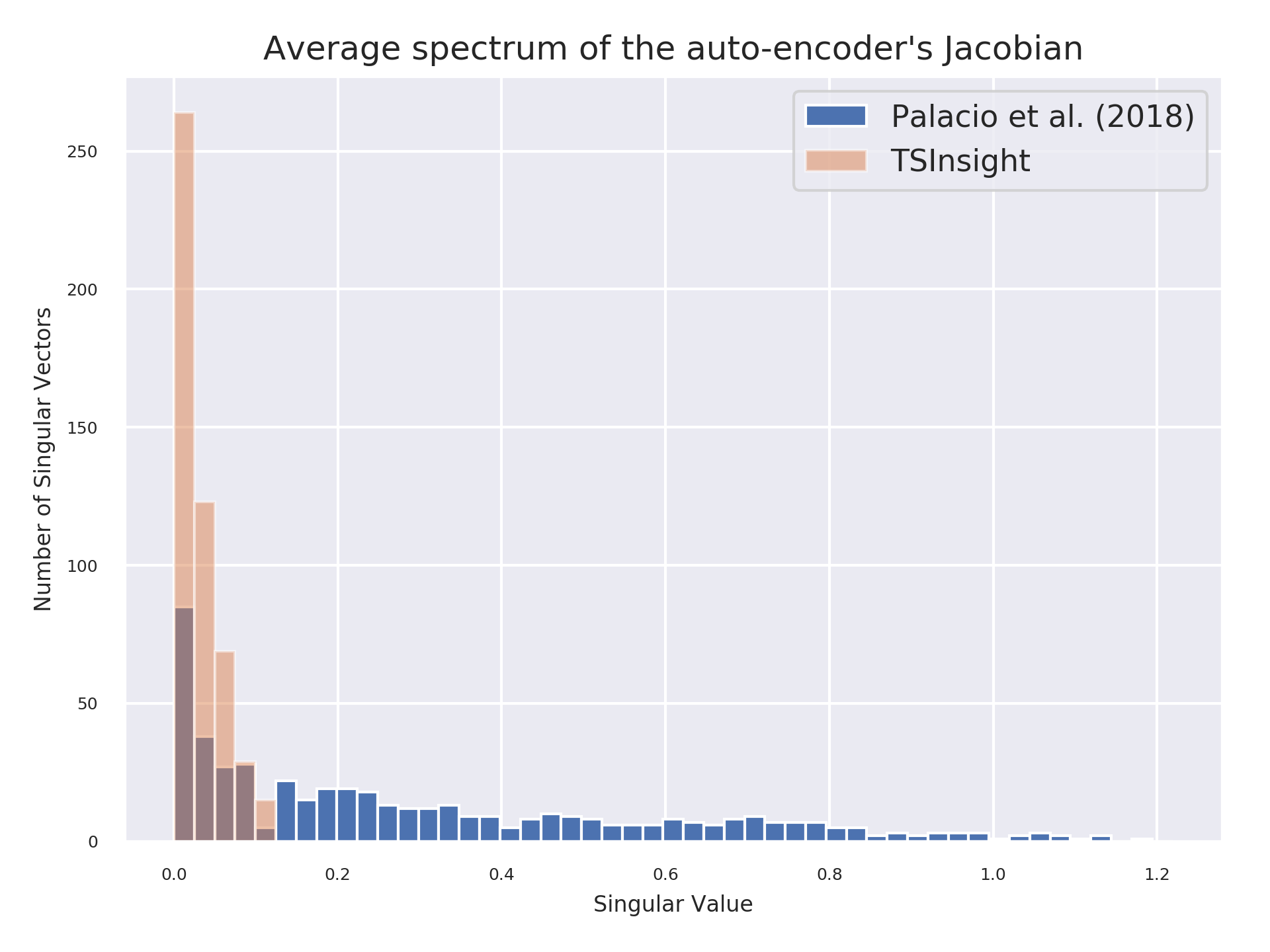}
    \caption{Spectrum analysis of the auto-encoder's average Jacobian computed over the entire test set of the forest cover dataset. The sharp decrease in the spectrum for TSInsight suggests that the network was successful in inducing a contraction of the input space.}
    \label{fig:spectrum_analysis}
    \vspace{-\intextsep}
\end{wrapfigure}


Fig.~\ref{fig:spectrum_analysis} visualizes the histogram of singular values of the average Jacobian on test set of the forest cover dataset. We compare the spectrum of the formulation from ~\cite{palacio2018deep} and TSInsight. It is evident from the figure that most of the singular values for TSInsight are close to zero, indicating a contraction being induced in those directions. This is similar to the contraction induced in contractive auto-encoders~\cite{rifai2011contractive} without explicitly regularizing the Jacobian of the encoder. 

\subsubsection{Generic Applicability}


\begin{figure*}[t]
    \centering
    \includegraphics[width=0.55\linewidth]{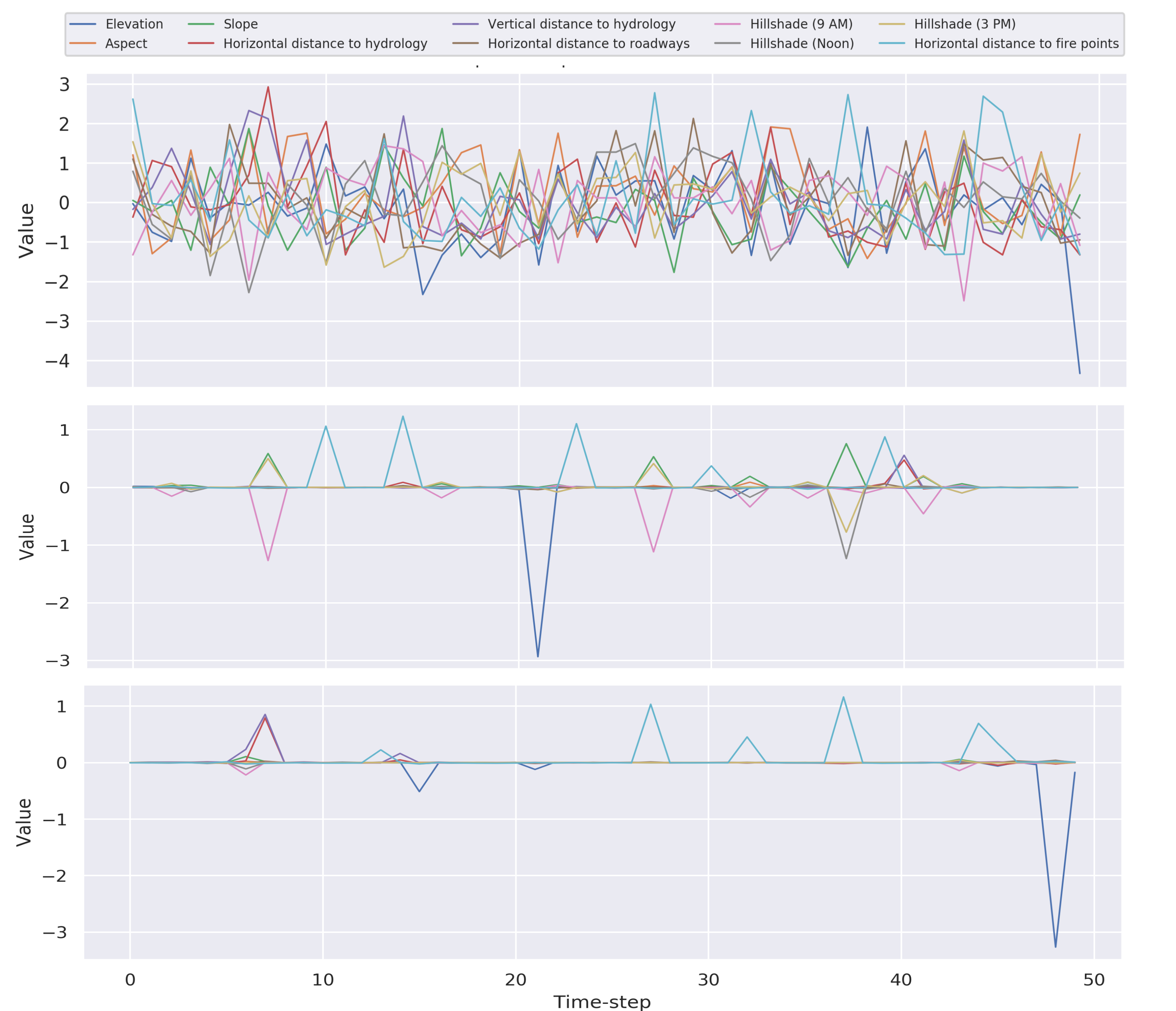}
    \caption{Auto-encoder training with different base models: (a) Raw signal (b) TSInsight attribution for CNN (c) TSInsight attribution for LSTM.}\label{fig:genericArch}
\end{figure*}


TSInsight is compatible with any base model. We tested our method with two prominent architectural choices in time-series data i.e. CNN and LSTM. 
The results highlight that TSInsight was capable of extracting the salient regions of the input regardless of the underlying architecture. 
It was interesting to note that since LSTM uses memory cells to remember past states, the last point was found to be the most salient. For CNN on the other hand, the network had access to the complete information resulting in equal distribution of the saliency. A visual example is presented in Fig~\ref{fig:genericArch}.

\section{Conclusion} \label{sec:conclusion}

We presented a novel method to discover the salient features of the input for the prediction by using the global context. With the obtained results, it is evident that the features highlighted by TSInsight are intelligible as well as reliable at the same time. In addition to interpretability, TSInsight also offers off-the-shelf properties which are desirable in a wide range of problems. Interpretability is essential in many domains, and we believe that our method opens up a new research direction for interpretability of deep models for time-series analysis.
We would like to further investigate the automated selection of hyperparameters in the future which is primitive for the wide-scale applicability of TSInsight along with its impact on adversarial robustness.


%
%
\bibliographystyle{splncs04}
\bibliography{ecml.bib}

\end{document}